\pdfoutput=1

\documentclass[11pt]{article}

\usepackage[final]{acl}
\usepackage{times}
\usepackage{latexsym}

\usepackage[T1]{fontenc}

\usepackage[utf8]{inputenc}

\usepackage{microtype}

\usepackage{inconsolata}

\usepackage{tabularx}
\usepackage{graphicx}
\usepackage{multirow}
\usepackage{tcolorbox}
\usepackage{xcolor} 
\usepackage{float} 
\usepackage{hyperref}
\usepackage{amsmath}  
\usepackage{amssymb}  
\usepackage{booktabs}
\title{A Transformer and Prototype-based Interpretable Model for Contextual Sarcasm Detection}

\author{Ximing Wen \\
  Drexel University, \\Philadelphia, USA \\
  \texttt{xw384@drexel.edu} \\\And
  Rezvaneh Rezapour \\
 Drexel University, \\Philadelphia, USA \\
  \texttt{sr3563@drexel.edu} \\}

\begin{document}
\maketitle
\begin{abstract}
Sarcasm detection, with its figurative nature, poses unique challenges for affective systems designed to perform sentiment analysis. While these systems typically perform well at identifying direct expressions of emotion, they struggle with sarcasm's inherent contradiction between literal and intended sentiment. Since transformer-based language models (LMs) are known for their efficient ability to capture contextual meanings, we propose a method that leverages LMs and prototype-based networks, enhanced by sentiment embeddings, to conduct interpretable sarcasm detection. Our approach is intrinsically interpretable without extra post-hoc interpretability techniques. We test our model on three public benchmark datasets and show that our model outperforms the current state-of-the-art. At the same time, the prototypical layer enhances the model's inherent interpretability by generating explanations through similar examples in the reference time. Furthermore, we demonstrate the effectiveness of incongruity loss in the ablation study, which we construct using sentiment prototypes.
\end{abstract}

\section{Introduction}
The task of automatically detecting sarcasm introduces a complex challenge in natural language processing (NLP). This nuanced task bridges the gap between sentiment analysis and text interpretation, highlighting the complexity of understanding and analyzing sarcasm in written language \citep{ilavarasan2020survey}. Sarcasm, characterized by a sharp, often humorous contrast between literal and intended meanings of statements, poses unique difficulties for computational models. These challenges stem from sarcasm's deep reliance on contextual clues, tone, and common human experiences. This complexity is further amplified in digital communication, where non-verbal cues are largely absent, making it essential to develop advanced models capable of interpreting such subtleties with high accuracy.

Deep learning models, especially transformer-based language models (LMs), have significantly contributed to advancements in NLP, offering powerful tools for sentiment analysis \citep{bu2024efficient,wang-dragut-2024-overlooked,Wang_Huang_Dragut_2026}, emotion detection \citep{tu2024adaptive}, and, by extension, sarcasm detection \citep{helal2024contextual}. 
More specifically, to study sarcasm detection, there is a trend to leverage LMs and a variety of features generated by different models to improve prediction accuracy \citep{cai-etal-2019-multi, bedi2021multi}. However, a persistent critique of deep learning models is their ``black-box'' nature, which obscures the decision-making process and hinders their interpretability. Current approaches usually adopt post-hoc interpretability methods Local Interpretable Model-agnostic Explanations (LIME) \citep{DBLP:journals/corr/RibeiroSG16}, and SHapley Additive exPlanations (SHAP) \citep{DBLP:journals/corr/LundbergL17} or attention mechanisms to explain a model's decision \citep{DBLP:journals/corr/RibeiroSG16,pmlr-v130-mardaoui21a,kumar2021explainable}. However, those explanations are still word-level and can only tell which part of the input they are looking at. As for sarcasm detection, when the text does not contain words that convey strong sentiment and instead uses other ways, such as analogy to express sarcasm, the word-level explanations could set similar weights to words in the sentence, and humans are ill-equipped to interpret them.

To address this challenge, our paper presents an intrinsically interpretable NLP framework that integrates prototype classification networks \citep{10.5555/3504035.3504467} with multi-view of semantic embedding and sentiment embedding from large-scale pre-trained transformer language models. To the best of our knowledge, our study is the first to apply a prototype-based network in sarcasm detection. This is achieved through a unique training regimen that enables the network to learn a collection of prototype tensors, which encapsulate latent clusters of training samples. At the point of inference, the model makes classification judgments solely based on the similarity to these prototypes, allowing for the model's decisions to be transparently explained by referencing the training examples most closely aligned with the top-matched prototypes. Together with a sentiment-prototype-based incongruity loss that captures the difference between implicit and explicit sentiment, our approach not only provides clear, human-understandable explanations for its predictions but also achieves state-of-the-art performance. 
The key contributions of our methods can be summarized as follows:
\begin{enumerate}
\item  We propose a novel interpretable framework for sarcasm detection. Our framework is built upon a prototype-based network leveraging semantic embedding and sentiment embedding from pre-trained transformer-based language models. 

\item  Extensive experiments on three public benchmark datasets show that our approach achieves state-of-the-art performance while being interpretable. We also conduct an ablation study to analyze the influence of incongruity loss in our model.

\item  We conduct case studies to show that our model can generate human-readable, sentence-level explanations for the model's reasoning process at the reference time. Our model and training code are available here: \url{https://github.com/social-nlp-lab/Sarcasm-Detection}.
\end{enumerate}

\section{Related Work}
\subsection{Contextual Sarcasm Detection}

Initial research in sarcasm detection primarily relied on simple lexical and syntactic features, and the classifiers are categorized as \textbf{Content-based Models} \citep{10.1145/1651461.1651471, davidov-etal-2010-semi,gonzalez-ibanez-etal-2011-identifying},leveraging features like n-grams, and part-of-speech tags \citep{riloff-etal-2013-sarcasm,Tepperman2006yeahRS,Tsur2010ICWSMA}. 

With the increase in the usage of sarcasm on online platforms in recent years, the performance of the sarcasm detection model is usually compromised in terms of robustness when faced with texts plagued by grammatical inaccuracies \citep{vsvelch2015excuse}. Moreover, these texts are usually a series of posts and comments that are highly temporal and contextual. As a result, relying solely on linguistic cues has become inadequate, prompting researchers to develop \textbf{Context-based Models}. A prominent strand of this research involves mining sentiment incongruity in sarcastic texts with attention-machenism \cite{pan2020modeling, najafabadi2024multi} to improve models' performance. Diverging from these methods, our approach adopts sentiment prototypes to discern both implicit and explicit sentiments within texts, enhancing the interpretability of the reasoning process. Furthermore, other scholars are exploring the modeling of user interactions via Graph Convolutional Networks (GCN) \cite{mohan2023sarcasm} or employing commonsense knowledge transformers like COMET \cite{yu2023commonsense}.



\subsection{Explainability of Transformer Language Models}
\paragraph{Post hoc Interpretability: }
When leveraging LM's ability to understand context, the model's complexity prevents people from understanding the model's reasoning process. In the field of XAI, post-hoc explainable approaches, such as LIME \citep{DBLP:journals/corr/RibeiroSG16}, and SHAP \citep{DBLP:journals/corr/LundbergL17}, are used to generate explanations for transformers' results \citep{article} through analyzing weights for each input word representation. However, these methods are now facing the challenge of being faithful and robust. Various studies show the sensitivity of the models to small perturbations in the target model's inputs or parameters \citep{ivankay2022fooling,pmlr-v130-mardaoui21a}. Other researchers, such as \citet{akula-garibay-2021-explainable} and \citet{attention2}, have explored the use of attention mechanisms to interpret prediction outcomes, focusing on how attention scores are allocated across individual words. However, these explanations remain at the word level, indicating only which parts of the input the model attends to. In the context of sarcasm detection, where sarcasm may be conveyed through mechanisms such as analogy rather than explicit sentiment, word-level explanations can assign similar weights to different words in a sentence. This makes it challenging for humans to interpret the underlying sarcastic intent effectively. To address this, we aim for more intuitive and sparse explanations: well-descriptive but short-sequence prototypes.

\paragraph{Prototype-based Reasoning in Deep Neural Networks: }
Prototype-based methods emphasize that visualizing the reasoning process through prototypes can significantly improve the intuitiveness of interpretation. This approach, leveraging prototype-based reasoning, has been a core aspect of interpretability in classical models for decades, as evidenced by research from \citet{10.1145/42411.42416}, \citet{10.1145/4284.4285}, and \citet{kim2015bayesian}. A pioneering example of incorporating prototypical learning into deep neural networks is the work by \citet{chen2019looks}, who introduced a novel neural network design for image classification. By inserting a prototype layer following the convolutional layers, the model compares convolution responses across different locations in the predicted image with predefined prototypes. Furthermore, this allows users to grasp why an image is classified in a certain way, such as understanding why a bird is identified as a `red-bellied woodpecker' due to its distinct red belly and head, along with black and white wing stripes. Following this work, researchers explored incorporating prototype layer with transformer-based encoders, such as Universal Sentence Encoder, BERT, BART (Bidirectional and Auto-Regressive Transformers) in fake news detection and hotel review classification \citep{das-etal-2022-prototex, 10.5555/3648699.3648963,wen2024language}. Sarcasm, due to its nature, can benefit from such reasoning provided by prototype-based models. However, this approach is still underexplored.

\begin{figure*}[t]
\centering
\includegraphics[width=1\linewidth]{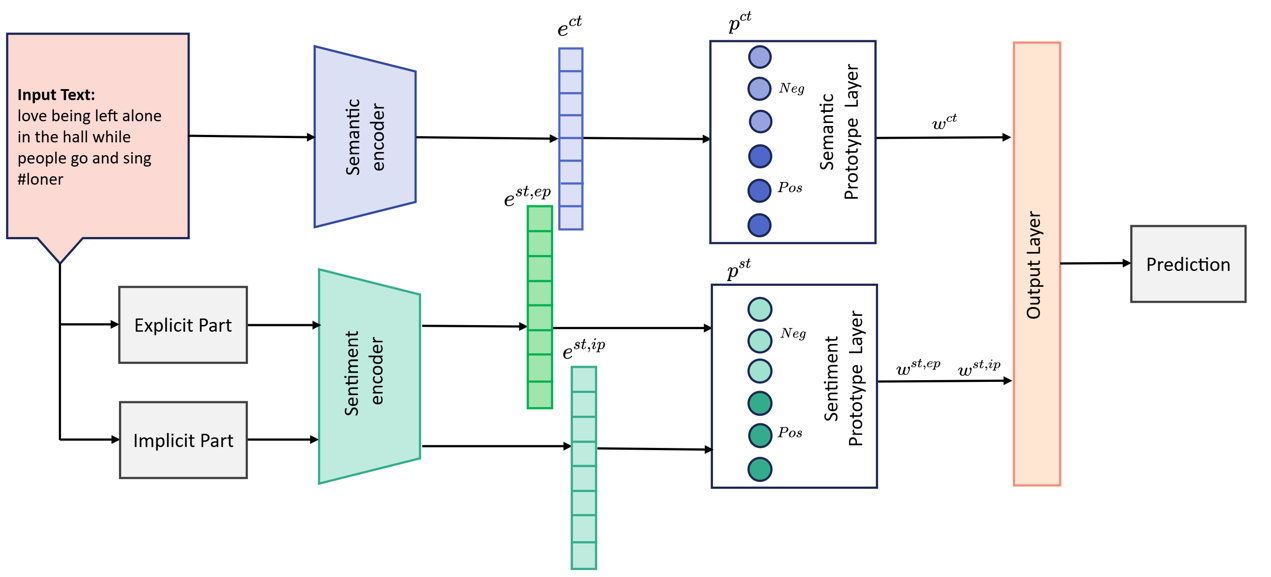} 
\caption{Diagram of the proposed model architecture and workflow.}
\label{fig:image} 
\end{figure*}
\section{Methodology}
Our approach, as shown in Figure \ref{fig:image}, first encodes semantic embedding and sentiment embedding with the Semantic View encoder and the Sentiment View encoder separately, and then both the encoded semantic embedding and sentiment embedding are fed into two separate prototype layers. Finally, the output distance vectors from the two prototype layers are concatenated and sent to the output layer to make the prediction.
\subsection{Semantic View}\label{sec:semantic}
The Semantic View focuses on capturing contextual meanings from text input. We use pre-trained language models (LMs) from Hugging Face \citep{wolf-etal-2020-transformers} as encoders to extract sentence-level embeddings. When the input comment has ancestor posts (\textit{e.g.}, SARC  Dataset by \citet{khodak-etal-2018-large}), which help set up the stage for the conversation and provide more context information, we concatenate the nearest ancestor with the comment as a whole and encode the embeddings with a pre-trained language model (\textit{e.g.} Sentence-BERT (SBERT) by \citealp{reimers2019sentencebert}, RoBERTa by \citealp{liu2019roberta}).

\begin{equation}
    \mathbf{e}^{ct} = \mathbf{Encoder(x)}
\end{equation}
where $x$ denotes the input text, $\mathbf{e^{ct}}$ is the semantic information representation vector.

\subsection{Sentiment View}
For sentiment feature extraction, we decompose the text into two parts: \textbf{Explicit Part}, which are sentiment words extracted following \cite{joshi2015harnessing}, and \textbf{Implicit Part}, which is the rest of the text capturing the implicit sentiments.

Both segments are processed through the Sentiment Encoder, specifically the SiEBERT model \cite{hartmann2023more}, to separately obtain the vector representations
$\mathbf{e}^{st,ep} $ and $ \mathbf{e}^{st,ip}$ of the CLS token from the final hidden state. For brevity, we use $ \mathbf{e}^{st}$ to represent both of them in the following equation:

\begin{equation}
    \mathbf{e}^{st} = \mathbf{SiEBERT(x)}
\end{equation}
where $\mathbf{x}$ denotes the input text, $\mathbf{e}^{st}$ is the sentiment information representation vector. The explicit representation \( \mathbf{e}^{st,ep} \) is labeled as $\mathbf{z}^{st,ep}$ with SiEBERT. The implicit representation \( \mathbf{e}^{st,ip} \) is labeled as $\mathbf{z}^{st,ip}$ identically to \( \mathbf{z}^{st,ep} \) for non-sarcastic inputs, whereas it is labeled oppositely for sarcastic inputs.

\paragraph{Fallback Strategy} The implementation incorporates a fallback strategy for cases where no explicit sentiment cues are identified in the input text. Specifically, when the sentiment word detector returns an empty set, a predefined neutral-to-positive sentiment token sequence is used as a surrogate explicit sentiment representation. This surrogate input is encoded using SiEBERT following the same preprocessing and inference pipeline applied to detected sentiment words, ensuring architectural consistency.

\subsection{Prototypical Layer}
As shown in Figure \ref{fig:image}, after the input is encoded into a latent semantic representation  $\mathbf{e}^{ct} \in \mathbb{R}^{d_s} $ through Semantic View and two latent sentiment representation $\mathbf{e}^{st,ep} \in \mathbb{R}^{d_m},\mathbf{e}^{st,ip} \in \mathbb{R}^{d_m}$ through Sentiment View separately, each representation is fed into a prototype layer respectively.

\paragraph{Semantic Prototype: } The Semantic Prototype Layer consists of $k_a$ prototypes $\mathbf{P}^{ct} = \{\mathbf{p}_j^{ct}\}_{j=1}^{k_a}$, where $\mathbf{p}^{ct}_j \in \mathbb{R}^{d_s} $, the same dimension of the encoded latent semantic feature  $\mathbf{e}^{ct}$. Each prototype, represented as a tensor, encapsulates a cluster of training examples. To ensure that both sarcastic and non-sarcastic classes are effectively represented by the learned prototypes, we allocate a fixed number of prototypes to each class. Utilizing k-means clustering \cite{hartigan1979algorithm}, we segment the training data of each class into multiple clusters and subsequently initialize the prototypes for each class with these cluster centers. These prototypes are trained through loss terms $\mathcal{L}_{cls}^{ct}$ and $\mathcal{L}_{sep}^{ct}$ defined in \S \ref{sec:loss}. This layer calculates the similarity between embedding $\mathbf{e}^{ct}$ and each prototype $\mathbf{p}_{j}^{ct}$ with a Radial basis function (RDF) kernel function as follows:
\begin{equation}
    sim(\mathbf{e}^{ct}, \mathbf{p}_j^{ct}) =exp\left(-\frac{\|\mathbf{e}^{ct} - \mathbf{p}_j^{ct}\|_2^2 + \varepsilon}{\sigma^2}\right)
\end{equation}

This similarity score increases monotonically as the Euclidean distance increases. $\sigma$ is a user-specified value that determines how quickly the similarity score increases as the distance between $\mathbf{e}^{ct}$ and $\mathbf{p}_j^{ct}$ decreases. A small $\sigma$ makes the kernel function more sensitive to changes in distance, leading to a kernel matrix with more localized information about the data points. This can make the model more sensitive to noise in the data. A large 
$\sigma$, on the other hand, produces a smoother kernel function that is less sensitive to the exact distance between data points, potentially making the model more robust. We also add a small value $\varepsilon$ set as $1e-4$ for numerical stability. We get vector $\mathbf{w}^{ct}$ by calculating the similarity score between $\mathbf{e}^{ct}$ and each prototype vector within $\mathbf{P}^{ct} $.
\paragraph{Sentiment Prototype Layer:}
The Sentiment Prototype Layer comprises $k_b$ prototypes $\mathbf{P}^{st} = \{\mathbf{p}_j^{st}\}_{j=1}^{k_b}$, where  $\mathbf{p}^{st}_j \in \mathbb{R}^{d_m}$. These prototypes are categorized into positive prototypes where $\mathbf{p}^{st} \in \mathbf{P}_{1}^{st}$ and negative prototypes where $\mathbf{p}^{st} \in \mathbf{P}_{0}^{st}$.
We initialize \( \mathbf{P}_{1}^{st} \) using the k-means cluster centers computed from the positive training data labeled by SiEBERT and \( \mathbf{P}_{0}^{st} \) using those derived from the negative training data. When clustering, we only use non-sarcastic training samples, without dividing the text into implicit and explicit parts.

We calculate the similarity between embedding $\mathbf{e}^{st}$ and each prototype $\mathbf{p}_j^{st}$ as follows:

\begin{equation}
    sim(\mathbf{e}^{st}, \mathbf{p}_j^{st}) =exp\left(-\frac{\| \mathbf{e}^{st}- \mathbf{p}_j^{st}\|_2^2 + \varepsilon}{\sigma^2}\right)
\end{equation}

We calculate similarity scores for the explicit and implicit representations, \( \mathbf{e}^{st, ep} \) and \( \mathbf{e}^{st, ip} \), against each prototype vector within \( \mathbf{P}^{st} \), yielding the similarity vectors \( \mathbf{w}^{st, ep} \) and \( \mathbf{w}^{st, ip} \), respectively.
\subsection{Output Layer}
The output layer is a fully connected layer followed by a sigmoid layer. It takes the concatenation of the extracted similarity vectors $\mathbf{w}^{ct}, \mathbf{w}^{st, ep} \text{ and } \mathbf{w}^{st, ip}$ from prototype layers as input and predict the likelihood of a text being sarcastic.

We intentionally employ concatenation to fuse these representations in order to preserve the independence of prototype similarity vectors, which is essential for interpretability in prototype-based models, as it allows direct attribution of the final prediction to individual semantic or sentiment prototypes. More complex fusion mechanisms, such as cross-attention or gating, may introduce stronger interactions but would entangle prototype signals and obscure their individual contributions. Interactions between explicit and implicit sentiment are therefore not modeled at the fusion layer; instead, they are explicitly captured by the incongruity loss (Eq. 12), which encourages disagreement between explicit and implicit sentiment predictions for sarcastic inputs.

\subsection{Loss and Training Algorithm}\label{sec:loss}
We construct the loss function with four different terms to ensure both accuracy and interpretability. 
\paragraph{Accuracy Loss: } The first term is accuracy loss, and it uses cross-entropy loss to optimize the predictive power of the network. In equation \ref{eq:prob}, $f$ is the output classifier, $\mathcal{P}_i$ is the predicted probability distribution, $y_i$ is the label, and $n$ is the total number of training data points. $\theta$ refers to the trainable weights in the classifier. 
\begin{equation}
   \mathcal{P}_i= f([\mathbf{w}^{ct}_i,\mathbf{w}^{st,ep}_i,\mathbf{w}^{st,ip}_i])
   \label{eq:prob}
\end{equation}
\begin{equation}
\mathcal{L}_{acc}   =  - \frac{1}{n}\sum_{i=1}^{n}\mathbf{y}_i\log P (\mathcal{P}_i=\mathbf{y}_i| \boldsymbol{\boldsymbol {\mathbf{x}_i} ; \theta})
\label{eq:sum1}
\end{equation}

\paragraph{Division Loss: } To distribute prototypes in the embedding space as much as possible, we design the $\mathcal{L}_{div}$ indicated in equation \ref{eq:sum2} for both semantic and sentiment prototypes. This loss uses cosine similarity to measure the difference between any two prototypes $\mathbf{p}_m$ and $\mathbf{p}_n$ in $P$ and penalizes it if their similarity is larger than $\lambda$. It is particularly beneficial when there are multiple prototypes for a single class, as it promotes the representation of diverse aspects of that class.

\begin{equation}
\mathcal{L}_{div}   = \sum_{\mathbf{p}_j, \mathbf{p}_q \in \mathbf{P},j\neq q}max(0,cos(\mathbf{p}_j, \mathbf{p}_q)- \lambda) 
\label{eq:sum2}
\end{equation}

\paragraph{Clustering and Seperation Loss: } The clustering loss $\mathcal{L}_{cls} $ and  $\mathcal{L}_{sep} $ are inspired by previous work, ProtoPNet \cite{chen2019looks}. The clustering loss $\mathcal{L}_{cls}$ ensures each embedding is close to at least one prototype in its own class, and separation loss $\mathcal{L}_{sep}$ encourages each embedding to be distant from prototypes not of its class. Together, $\mathcal{L}_{cls} $ and  $\mathcal{L}_{sep}$ push each prototype to focus more on training examples from the same class and less on training examples from other classes. 

For \textbf{semantic prototypes}, they are defined as follows:

\begin{equation}
\mathcal{L}_{cls}^{ct}   = \frac{1}{n}\sum_{i=1}^{n}\min_{\mathbf{p}_j^{ct} \in \mathbf{P}_{y_i}^{ct}} \|\mathbf{e}^{ct}_i-\mathbf{p}_j^{ct}\|_2^2 
\label{eq:sum3}
\end{equation}

\begin{equation}
\mathcal{L}_{sep}^{ct}   = - \frac{1}{n}\sum_{i=1}^{n}\min_{\mathbf{p}_j^{ct} \notin \mathbf{P}_{y_i}^{ct}} \|\mathbf{e}^{ct}_i-\mathbf{p}_j^{ct}\|_2^2
\label{eq:sum4}
\end{equation}

For \textbf{sentiment prototypes}, we promote proximity between positive prototypes and training data segments labeled positive ($z_i=1$) by SiEBERT, and likewise align negative prototypes with training data segments labeled negative ($z_i=0$) through losses defined as follows:

\begin{align}
\mathcal{L}_{cls}^{st} &= \frac{1}{n}\sum_{i=1}^{n}\min_{\mathbf{p}_j^{st} \in \mathbf{P}_{z_i}^{st}} \|\mathbf{e}^{st}_i-\mathbf{p}_j^{st}\|_2^2 \label{eq:sum5} \\
\mathcal{L}_{sep}^{st} &= - \frac{1}{n}\sum_{i=1}^{n}\min_{\mathbf{p}_j^{st} \notin \mathbf{P}_{z_i}^{st}} \|\mathbf{e}^{st}_i-\mathbf{p}_j^{st}\|_2^2 \label{eq:sum6}
\end{align}

The final Clustering and Separation Loss is $\mathcal{L}_{cls\_sep}=\mathcal{L}_{cls}^{ct}+\mathcal{L}_{sep}^{ct}+\mathcal{L}_{cls}^{st}+\mathcal{L}_{sep}^{st}$

\paragraph{Incongruity Loss: } We hypothesize the presence of incongruity between the explicit and implicit sentiments within a sarcastic comment. For instance, in the sarcastically labeled sentence, ``Oh no, a rainy day again! This is great!'' the explicit sentiment conveyed by the word ``great'' appears positive. However, upon closer examination of the context, it becomes evident that the speaker does not favor rainy days, revealing an underlying negative sentiment. Based on this observation, we introduce the incongruity loss defined with cross-entropy to effectively capture this disparity between explicit and implicit sentiment:

\begin{equation}
\begin{split}
\mathcal{L}_{inco} = & - \frac{1}{n}\sum_{i=1}^{n} \left( \mathbf{z}_i^{ep}\log P (h(\mathbf{w}_i^{st,ep})=\mathbf{z}_i^{ep}| \boldsymbol{\theta}) \right. \\
                     & \left. + \mathbf{z}_i^{ip}\log P (h(\mathbf{w}_i^{st,ip})=\mathbf{z}_i^{ip}| \boldsymbol{\theta}) \right)
\label{eq:inco}
\end{split}
\end{equation}



where $h$ is a MLP based classifier that predicts a probability distribution. 

We construct the final loss function \( L \) by combining the previously defined loss components, each weighted by their respective coefficient \( \lambda \). Additionally, we incorporate an L1 regularization term \( \|\theta\| \) to promote sparsity in the weights of the output layer. The final loss $\mathcal{L}$ is defined as follows:

\begin{equation}
\mathcal{L} =  \mathcal{L}_{acc}  +\lambda_1\mathcal{L}_{div} +\lambda_2\mathcal{L}_{cls\_sep} +\lambda_3\mathcal{L}_{inco} +\lambda_4\|\theta\|
\label{eq:sum4}
\end{equation}

\subsection{Prototype Projection}
For improved interpretability, we visualize the semantic prototypes by projecting a prototype vector onto its closest datapoint in the training dataset, measured by Euclidean distance. For the large training dataset with over 120k comments, we did a random pre-sampling of one-tenth of the comments for computation efficiency. Each prototype's embedding is replaced with the nearest comment's embedding in the training data point. The alignment of prototypes with training set samples provides an intuitive and easily understandable interpretation for humans.

\subsection{Prototype Initialization}
We use k-means cluster centers on the training data to initialize prototypes, with a fixed small number of prototypes per class (20). We observed the model to be robust to prototype counts within a reasonable range; extremely small counts reduce coverage, and extremely large counts increase redundancy but yield similar performance due to the division loss.

\section{Experimental Setup}\label{sec:experiment}
In this section, we introduce the datasets used in the experiment and the baseline models.
\subsection{Data}
We evaluate our methods on the following three public benchmark datasets: (1) \textbf{SARC 2.0} \footnote{Link: \href{ https://nlp.cs.princeton.edu/old/SARC/2.0/} {https://nlp.cs.princeton.edu/old/SARC/2.0/}} \citep{khodak-etal-2018-large}, a corpus comprising 1.3 million comments on Reddit. Each comment is self-annotated, and we focus on the primary main balanced variation of it, with 118,940 comments in training and 56,118 comments in the testing set. (2) \textbf{Twitter} \citep{riloff-etal-2013-sarcasm} dataset is sourced from the Twitter platform. Sarcastic tweets are identified using the hashtag \#sarcasm, while non-sarcastic tweets lack this hashtag. The dataset consists of 1,368 training examples and 588 test examples. (3) \textbf{Sarcasm Corpus V2 Dialogues} \citep{oraby2017creating} is a diverse and richly annotated corpus of sarcasm in dialogue. This dataset is collected from a variety of dialogue sources to capture sarcasm in different conversational contexts, moving beyond traditional social media platforms like Twitter.

\subsection{Models and Settings}
We employed 5-fold cross-validation to evaluate our model's performance and fine-tuned the hyperparameters on the validation data. The Optimizer for all neural networks is Adam, and the learning rate is $1e-4$. We used one single GTX 3090 for each model's training, and due to the limitation of GPU RAM, when training with LM encoders, we chose a batch size of 60 and an accumulated gradient step of 30. We use early stopping \citep{pytorch-ignite} based on the loss of validation data.

\paragraph{Semantic Encoder: }
In prototype-based architectures, it is a common practice to use Euclidean distance to measure sentence similarity between training examples and prototypes. However, embeddings from transformer-based language models are not typically trained with contrastive loss that leverages Euclidean distance for measuring sentence similarity. To further explore whether this influences the model's interpretability, we selected RoBERTa-large and SBERT as the encoders for comparison in our experiments:

\paragraph{\indent (1) RoBERTa-large: } 
RoBERTa-large is a transformer-based language model developed to improve upon the BERT architecture. We used the [CLS] token embedding from the last hidden states directly for the downstream task.

\paragraph{\indent (2) SBERT: } 
SBERT is post-trained on  BERT. It uses siamese and triplet network structures to derive semantically meaningful sentence embeddings that can be compared using cosine similarity. We used the pre-trained model all-mpnet-base-v2 \footnote{The model can be downloaded here: \href{https://huggingface.co/sentence-transformers/all-mpnet-base-v2}{https://huggingface.co/sentence-transformers/all-mpnet-base-v2}} developed by Microsoft from Hugging Face.

\paragraph{Sentiment Encoder: }
For sentiment feature extraction, we employed the SiEBERT model \cite{hartmann2023more}, which is a fine-tuned version of the RoBERTa-large model optimized for sentiment classification tasks. 
\paragraph{Baseline}: 
For comparative evaluation with our approach, we selected the following methodologies as baselines: Fracking Sarcasm \cite{ghosh2016fracking}, GRNN \cite{zhang2016tweet}, CNN-LSTM-DNN \cite{ghosh2016fracking}, SIRAN \cite{tay2018reasoning}, MIRAN \cite{tay2018reasoning}, ELMo-BiLSTM \cite{ilic2018deep}, A2Text-Net \cite{liu2019a2text}, SARC 2.0 \cite{khodak-etal-2018-large}, and CASCADE \cite{C18-1156}. Additionally, we incorporated GRU-Attention model \cite{akula-garibay-2021-explainable}, which offers intrinsic interpretability through attention scores. We further considered a graph-based method, BERT-GCN \cite{mohan2023sarcasm}, alongside ensemble strategies such as Fuzzy-Logic \cite{dai2024bert} and MULE \cite{vitman2023sarcasm}. BiGRU \cite{najafabadi2024multi} is included for its ability to measure sentiment incongruity.

\paragraph{Metrics: } We used accuracy, recall, and F1-Score as metrics to evaluate models' performance. Since we use 5-fold cross-validation, we calculated the average of 5 experiment results on the test dataset for each metric as the final result.


\begin{figure*}[t]
\centering
\includegraphics[width=0.97\textwidth]{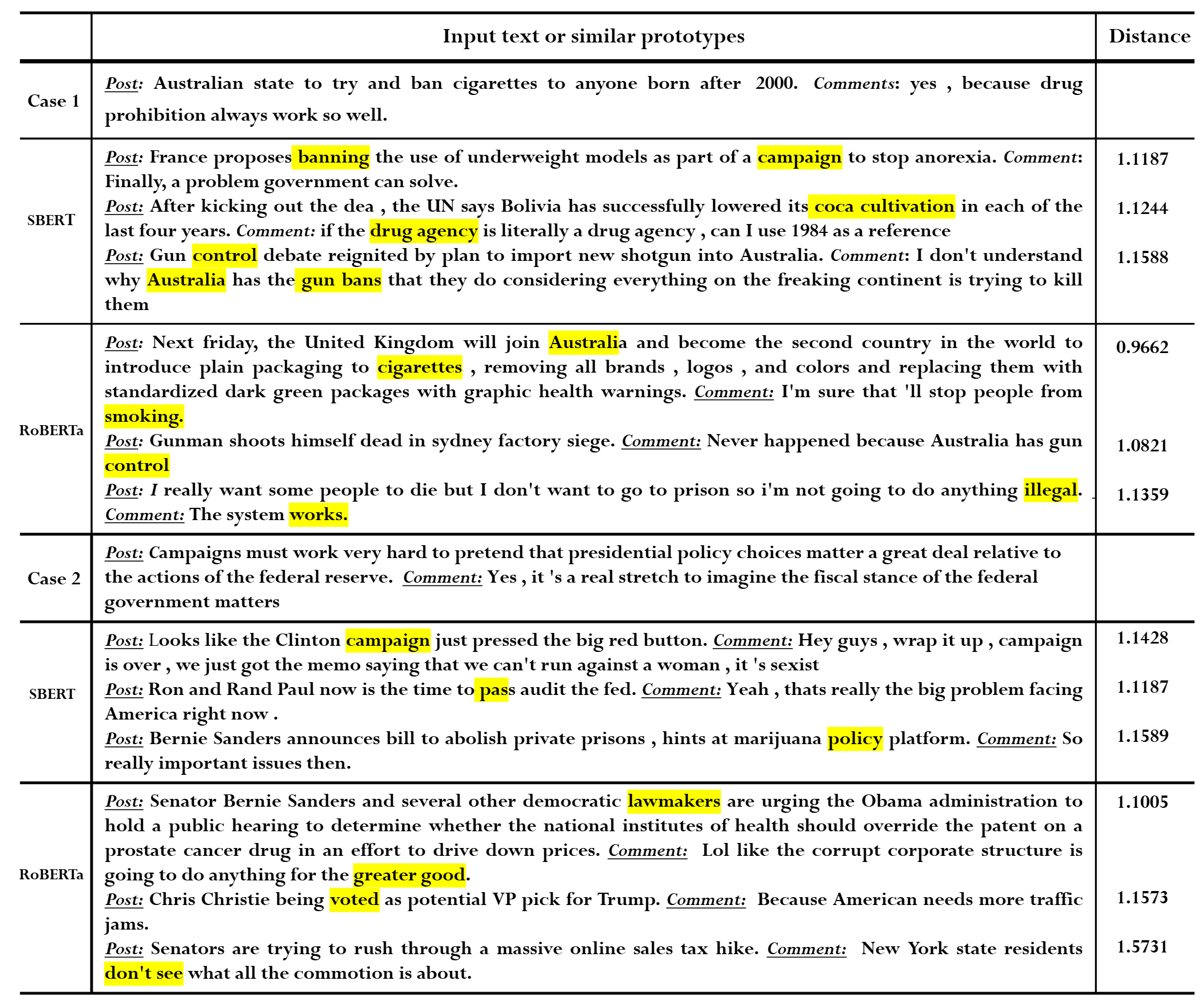} 
\captionof{table}{Case examples of input texts, semantic prototypes of SBERT and RoBERTa after projection, and the Euclidean distance score between the prototype embeddings and input sentence vector. \textbf{Note:} The highlighted words are only meant for reader guidance and were manually annotated by us to aid comparison.}
\label{fig:case} 
\end{figure*}
\section{Results \& Discussion}
We mainly focus on answering the two questions through our experiments: \textbf{Q1:} What is the performance of our white box framework compared to other baselines, and \textbf{Q2:} How well does the given explanation represent the true reasoning of the model?  We discuss them separately in the following sections.

\begin{table}[h!]
\small
\centering
\begin{tabular}{lccc}
\hline
\textbf{Models} & \textbf{Accuracy} & \textbf{Recall} & \textbf{F1} \\
\hline
Fracking Sarcasm & 89.2 & 87.9 & 88.1 \\
GRNN  & 66.4 & 64.7 & 65.4 \\
ELMo-BiLSTM  & 76.2 & 75.0 & 75.9 \\
ELMo-BiLSTM FULL  & 77.4 & 73.5 & 75.3 \\
ELMo-BiLSTM AUG  & 68.6 & 70.8 & 69.4 \\
A2Text-Net & 91.3 & 91.0 & 90.0 \\
GRU-Attention  & 97.9 & \textbf{99.6} & 98.7 \\
BERT-GCN & 88.3 & 87.1&  87.3 \\
MULE & 93.5  & 94.1 & 93.8 \\
\hline

\textbf{Our Model} (SBERT)& 98.0 & 97.3 & \textbf{98.7} \\
\textbf{Our Model} (RoBERTa) & \textbf{98.3} & 98.6 & 98.4 \\
\hline
\end{tabular}
\caption{Results on Twitter dataset.}
\label{res1}
\end{table}

\begin{table}[h!]
\small
\centering
\begin{tabular}{lccc}
\hline
\textbf{Models} & \textbf{Accuracy} & \textbf{Recall} & \textbf{F1} \\
\hline
GRNN  & 64.4 & 61.8 & 61.2 \\
CNN-LSTM-DNN  & 67.3 & 66.7 & 65.7 \\
SIARN & 71.9 & 71.8 & 71.8 \\
MIARN & 74.2 & 72.9 & 72.7 \\
ELMo-BiLSTM  & 74.6 & 74.7 & 74.7 \\
ELMo-BiLSTM FULL  & 76.2 & 76.0 & 76.0 \\
GRU-Attention  & 77.3 & 77.2 & 77.2 \\
BiGRU & 79.3 & 81.4 &80.2  \\
Fuzzy-Logic & 81.8  & 80.3 & 81.0 \\

\hline
\textbf{Our Model} (SBERT) & 82.1 & 82.5 & 82.4 \\
\textbf{Our Model} (RoBERTa) & \textbf{83.6} & \textbf{83.7} & \textbf{83.6} \\

\hline
\end{tabular}
\caption{Results on Sarcasm Corpus V2 Dialogues dataset}
\label{res2}
\end{table}

\begin{table*}[h!]
\centering
\footnotesize
\begin{tabular}{lccc cccc}
\hline
 & \multicolumn{3}{c}{\textbf{w}} & \multicolumn{3}{c}{\textbf{w/o}} \\
\cmidrule(lr){2-4} \cmidrule(lr){5-7}
 & \textbf{Acc.} & \textbf{Rec.} & \textbf{F1} & \textbf{Acc.} & \textbf{Rec.} & \textbf{F1} \\
\hline
Twitter & 98.3 (+1.1\%) & 98.6 (-0.1\%) & 98.4 (+0.9\%)& 97.2 & 98.7 & 97.5 \\
Dialogues & 83.6 (+1.7\%) & 83.7(+1.3\%) & 83.6 (+1.3\%) & 82.2 & 82.6 & 82.5 \\
SARC & 82.4 (+2.2\%) & 85.8 (+2.4\%) & 83.0 (+2.9\%) & 80.2 & 83.4 & 81.1 \\
\hline
\end{tabular}

\caption{Ablation study on the effect of incongruity loss, where `w' denotes models trained with incongruity loss, and `w/o' refers to models trained without it. The percentage reflects the variation in performance when training with incongruity loss compared to training without it.}
\label{inco}
\end{table*}

\begin{table}[h!]
\small
\centering
\begin{tabular}{lccc}
\hline
\textbf{Models} & \textbf{Accuracy} & \textbf{Recall} & \textbf{F1} \\
\hline
CASCADE  & 77.0 & 84.0 & 77.0 \\
SARC 2.0 & 75.0 & - & 76.0 \\
ELMo-BiLSTM  & 72.0 & - & -\\
ELMo-BiLSTM FULL  & 76.0 & & 76.0 \\
GRU-Attention  & 81.0 & 82.1 & 81.0 \\
BiGRU & 69.4 & 68.6 &69.0  \\
MULE & 75.2  &83.8 & 80.1 \\
\hline

\textbf{Our Model} (SBERT)& 80.1 & \textbf{86.2} & 82.2 \\
\textbf{Our Model} (RoBERTa) & \textbf{82.4} & 85.8 & \textbf{83.0} \\
\hline
\end{tabular}
\caption{Results on Reddit dataset SARC 2.0}

\label{res3}
\end{table}

\begin{table}[h]
\small
\centering
\begin{tabular}{lccc}
\hline
\textbf{Method} & \textbf{Twitter} & \textbf{Dialogues} & \textbf{SARC} \\
\hline
$E_{ct}$ & 97.3 & 82.7 & 81.02 \\
$E_{ct} + E_{st}$ (w/o $L_{inco}$) & 97.5 & 82.5 & 81.1 \\
$E_{ct} + E_{st}$ (with $L_{inco}$) & 98.4 & 83.6 & 83.0 \\
\hline
\end{tabular}
\caption{Ablation study on different modules}
\label{tab:ablation}
\end{table}

\subsection{Overall Performance}
We represent our experiments on three public benchmark datasets in Table \ref{res1}, Table \ref{res2}, and Table \ref{res3} separately. 
Overall, our model achieved the highest accuracy, recall, and F1 scores on the Sarcasm Corpus V2 and SARC datasets, outperforming all other baselines. On the Twitter dataset, our model obtained the best accuracy and F1, while GRU-Attention yielded the best recall.
Additionally, to compare our approach with CASCADE on the SARC dataset, we incorporated personality features into our model. Of the two versions of our model, the one using a RoBERTa encoder generally outperformed its SBERT-based counterpart.

\subsection{Case Study for Explanations}
 
In Table \ref{fig:case}, we present trained semantic prototypes after projection and the distance score between prototypes and the input comment. The prototypes exhibit a similar topic as the input text, and the keywords are highlighted in yellow by humans for easy illustration (The highlighted words in Table \ref{fig:case} were only meant for reader guidance to aid comparison). For instance, in Case 1, the SBERT prototypes consist of comments skeptical of the efficacy of specific government regulations on cigarettes, and drugs, showing significant overlap with the input comments—a pattern that is also observed in Case 2. Our case study demonstrates that prototype-based models provide more intuitive and human-readable explanations for sarcasm than analyzing a distribution of scores, especially in the absence of strong sarcasm cue words.

When comparing the explanations generated by SBERT and RoBERTa, we did not see a significant difference between them, indicating that although, unlike SBERT, RoBERTa is not additionally trained to represent semantic similarity, it still works well with the prototype structure.

\subsection{Ablation Study}
To fully evaluate our model, we conducted an ablation study evaluating the influence of our \textbf{Incongruity Loss}. We experimented with and without the incongruity loss with the RoBERTa encoder on all three datasets and the result is shown in Table \ref{inco}.

Upon analyzing the results for the Dialogues and SARC datasets, we observed performance improvements ranging from -0.1\% to 2.9\% when training model with incongruity loss, with SARC dataset showing a particularly notable increase of over 2\% on recall, accuracy, and F1. In contrast, the Twitter dataset demonstrated minimal improvement. We hypothesize that this is due to the already high baseline performance on the Twitter dataset, which constrains the extent of further performance gains.

We further perform an ablation study to examine the effect of incorporating the sentiment encoder module in addition to the semantic encoder. The corresponding F1 scores are presented in Table~\ref{tab:ablation}. Without the incongruity loss, adding the sentiment encoder yields a comparable F1 score to using only the semantic encoder. However, when the incongruity loss is introduced, the model’s performance improves substantially, achieving up to a 2.9\% increase in F1.

\section{Conclusion}
We proposed a novel approach that leverages state-of-the-art LM encoders and prototype-based networks to build an intrinsically interpretable model for sarcasm detection. Our approach achieved state-of-the-art performance on three public benchmark datasets. By representing prototypes with the closest training sentences, our method can explain sarcasm detection with sentence-level, human-readable explanations. 
\section{Limitation}
Our data primarily comes from English-speaking populations from specific platforms, which may not be generalizable. Also, our work does not provide examples to show reasoning with sentiment prototypes. Future research could investigate generating explanations by analyzing the attention scores associated with both negative and positive sentiment prototypes.

\section{Ethics Statement}
The development and deployment of sarcasm detection models present several ethical considerations. First, our study recognizes the potential biases inherent in training data, particularly those that stem from subjective interpretations of sarcasm across different linguistic, cultural, and social contexts. We used three different datasets from diverse platforms to ensure generalizability.
In addition, while our prototype-based explainability framework enhances interpretability, we stress that sarcasm detection remains an inherently complex and nuanced task. We encourage the responsible use of our model to enhance human understanding rather than replace human judgment in sensitive contexts.
Therefore, we do not recommend over-reliance on automated sarcasm detection in complex tasks and advocate for human oversight in critical decision-making processes.
Moreover, we are committed to transparency in our research by making our model and code publicly available to foster reproducibility. We also adhere to ethical guidelines and ensure that all datasets used in this study comply with proper licensing.

\bibliography{custom}

\appendix



\end{document}